\documentclass[11pt,a4paper]{article}
\usepackage{acl}

\usepackage{times}
\usepackage{latexsym}
\usepackage[T1]{fontenc}
\usepackage[utf8]{inputenc}
\usepackage{microtype}
\usepackage{inconsolata}
\usepackage{graphicx}
\usepackage{subcaption}
\usepackage{relsize}

\usepackage{enumitem}
\setlist[itemize]{noitemsep, topsep=0pt}
\setlist[enumerate]{noitemsep, topsep=0pt}

\usepackage{tabularx}
\usepackage{booktabs}
\usepackage[multiple]{footmisc}

\title{On Tables with Numbers, with Numbers}

\author{
    Konstantinos Kogkalidis \\
    Aalto University \\
    Department of Computer Science \\
    \texttt{kokos.kogkalidis@aalto.fi}
    \And
    Stergios Chatzikyriakidis \\
    University of Crete \\
    Department of Philology \\
    \texttt{stergios.chatzikyriakidis@uoc.gr}
}

\newcommand{\eg}{\textit{e.g.},}
\newcommand{\ie}{\textit{i.e.},}
\newcommand{\ia}{\textit{inter alia}}
\newcommand{\etc}{\textit{etc.}}

\date{}

\begin{document}

\maketitle

\begin{abstract}
    This paper is a critical reflection on the epistemic culture of contemporary computational linguistics, framed in the context of its growing obsession with tables with numbers. We argue against tables with numbers on the basis of their epistemic irrelevance, their environmental impact, their role in enabling and exacerbating social inequalities, and their deep ties to commercial applications and profit-driven research. We substantiate our arguments with empirical evidence drawn from a meta-analysis of computational linguistics research over the last decade.
\end{abstract}

\section{Introduction}
Throughout its evolution, computational linguistics has undergone multiple identity crises.
In its present from, and despite its logical origins and linguistic ambitions, it is almost entirely aligned with positivist principles and ideals~\cite{church2021future}.
The imprint of this alignment is an idealization of experimental quantification, most commonly manifesting in the form of \textit{tables with numbers}.
Table with numbers can certainly be useful. 
That said, their centrality in contemporary computational linguistics borders on a pathological mixture of scientific reductionism and technological obsession.
Beneath the numbers lie signs of a field in disarray: a waning reliance on theory (linguistic or otherwise), nowadays substituted by model scale; a disproportionate representation of big industry and big academia, in turn associated with a lack of transparency, accessibility and inclusion; an experimental paradigm dominated by stagnant "tasks" and benchmarking practices, detached from technical rigor as well as scientific insight; and a progressive estrangement from societal, humanistic and environmental context.
And while the community seems to be both alert to and uneasy with the current state of affairs~\cite{michael-etal-2023-nlp}, a holistic analysis of these issues has been long missing from the literature.

In this paper, we brave a look under the number rock.
We conduct a critical assessment of the epistemic culture of computational linguistics, focusing specifically on its relation to tables with numbers.
We narrow down on four axes of interest:
\begin{itemize}
    \item The epistemological preconditions that granted tables with numbers the status of scientific currency, and the mechanisms that affect their actual value (\S\ref{sec:epistemological}).
    \item Their environmental footprint and the normative discourse around it (\S\ref{sec:environmental}).
    \item Their cause-and-effect relation to the perpetuation and exacerbation of inequality and harmful power structures (\S\ref{sec:societal}).
    \item Their intrinsic ties with corporate interest, profit, and the accumulation of technoscientific capital (\S\ref{sec:monetary}).
\end{itemize}

\begin{figure}[t]
    \centering
    \includegraphics[clip, trim=0.5cm 0 0 0,width=0.49\textwidth]{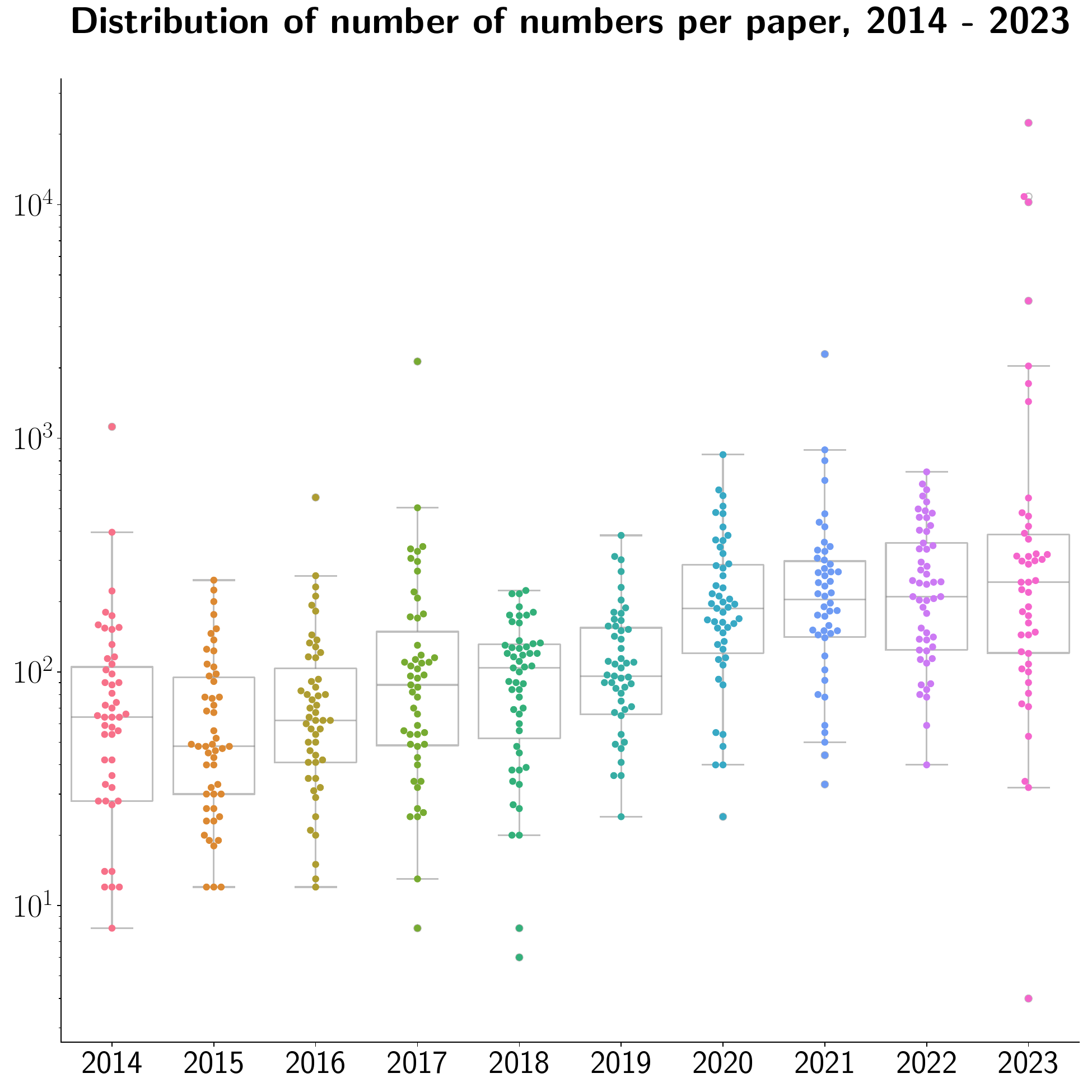}
    \caption{Box- and swarm-plots of the distribution of the number of experimental results per paper, grouped by year. We manually count the number of numbers within tables from the 50 most cited papers per year. We do not include numbers that pertain to descriptive dataset statistics, nor numbers reporting dispersion statistics (\eg{} confidence intervals, standard deviations \etc{}). The pattern indicates a marked upwards trend over time. Most (75\%) contemporary papers contain 100 to 300 numbers, while some (25\%) contain up to 1\,000.}
    \label{fig:numbers}
\end{figure}

\section{The Multiple Facets of Number}\label{sec:epistemological}
The field's dominant scientific approach embodies a wildly exaggerated version of positivism.
This is evident both in the themes prevalent in the mainstream discourse, and in those notably absent from it.
Two critical perspectives arise.
First, how faithfully does computational linguistics \textit{actually} adhere to its positivist posture? 
And second, what are the \textit{implications} of computational linguistics as a singularly positivist discipline?
We begin by addressing the former, setting off with a simplified introduction to the positivist worldview and its tenets.

\subsection{Number as Virtue}
As a scientific meta-theory, positivism asserts that knowledge is the yield of systematic, unbiased and reproducible observation.
A prospective theory is evaluated based on how well it can predict and interpret observations.
An impartial and irrefutable evaluation is what ensures theories can be refuted and reliably compared.
Ultimately, the essence of scientific progress lies in the iterative process of theory testing, rejection, and refinement.
This worldview holds truth as objective and unique, asserted as such by reproducibility, generalization, neutrality, and universality~\cite{ayer1959logical}.
Tables with numbers attain epistemic significance in bearing witness to this (idealization of) truth.

\subsection{Number as Number}
Alas, linguistic theories have fallen short of historical expectations.
To date, there is no hint of a consensus on what a concretely implementable mechanization of human language should (or even could) look like.
In lieu of theories, computational linguistics had to turn to the next best thing: models.%
\footnote{This is one reading. Another reading is that when machine learning ``solved'' vision, it moved over to NLP, setting aside linguistic expertise to make room for all the luggage it brought with it.}\textsuperscript{,}%
\footnote{The modern tendency to look for a theory \textit{within} the model is further evidencing the poverty of historical theories; see \citet{baroni2022proper,piantadosi2023modern}, \ia{}.}
Models promise less but do more, prioritizing tangible solutions over abstract notions of inquisitive deduction.
Apart from this deviation, the positivist methodological narrative is easy to recognize in the field's experimental pipeline.
Large datasets are heralded as authoritative collections of empirical observations, systematically condensing linguistic truth.
Datasets enact ``benchmarks'', standardized and fair test suites through which we can ``track progress'', \ie{} decide whether a model advances science, and if so, by \textit{how much}.
Congruent with the literature's makeup over the last decade, this suggests that contributions may come in one of two primary forms: models and benchmarks, dual facets of one and the same thing -- tables with numbers.

Nonetheless, in having discarded theory, the model-and-benchmark pipeline fails to uphold the scientific promise upon which it was built.
A first problem lies in the fact that the models developed and adopted nowadays are almost exclusively generic and theory-neutral~\cite{sutton2019bitter}.
In making no assumptions and yielding no hypotheses over their domain, they are infallible in all aspects except for their performance~\cite{schlangen-2021-targeting}.
The side effect is that the field's progress translates to technical know-how rather than an advance in the sum total of ``pure'' knowledge~\cite{krenn2022scientific,messeri2024artificial}.
Other than modeling insights, nothing gets in and nothing gets out, confining a traditionally interdisciplinary endeavour to a technocratic and opinionless monoculture.

A second, perhaps bigger, problem lies in the reductionist view of language faculty as something that can be broken apart into high-level ``tasks'', at the intersection of which one can find, and therefore \textit{quantify}, ``understanding''~\cite{raji2021ai}.
The verity of this assumption is not immediately obvious; modern models breeze through benchmarks, yet we remain as far as ever from attaining a holistic and comprehensive computational account of language.
The picture is sufficiently clear: sidetracked by models and benchmarks, computational linguistics has turned into natural language processing: a domain-specific engineering discipline that answers more questions than it asks.

\subsection{Number as Nothing}
Ironically, the remarkable ease of model iteration (as compared to the painstakingly slow process of theory iteration) is an inflationary factor for the epistemic value of numbers.
When experimental superiority becomes a prerequisite to publication~\cite{Rogers_2020_reviewing-models}, all publications invariably achieve it, rendering both the message (experimental superiority) and the messenger (publications) meaningless.
Immediate, short-sighted gains dominate the research agenda, and difficult questions become eschewed for the sake of incremental tweaks and micro-improvements~\cite{bhattacharya2020stagnation}.
Short-sighted goals are echoed in short-term memory, leading to plentiful instances of knowledge recycling, paper duplication and citation amnesia~\cite{singh2023forgotten}.
The over-standardization of form gradually turns into an equilibrium of intent -- contributions are pushed towards structural and semantic uniformity, ending up virtually indistinguishable from one another.
The frantic pace of ``progress'' turns scientific enterprise into a competition for experimental superiority, eroding integrity and transparency.
The most successful models are too time- and resource-consuming to replicate and cross-validate, leading to statistically insignificant tables filled with under-sampled and noisy numbers of dubious quality and utility~\cite{dodge-etal-2019-show,ethayarajh2020utility,belz-etal-2021-systematic}.
Scientific communication espouses sales pitch aesthetics, exaggerating merit, obscuring weakness and purposefully avoiding critical self-reflection and honest self-assessment~\cite{smaldino2016natural,lipton2019troubling}.
After a bountiful decade of benchmarking frenzy, there is now growing consensus that datasets are statistically biased, annotation is subjective, models are sensitive to noise -- the numbers have been \textit{lying all along}~\cite[\ia]{mccoy2019right,geirhos2020shortcut,liao2021we,plank2022problem}.%
\footnote{The fact that benchmarking is being made obsolete by a handful of closed source  models far beyond the community's reach is clearly just a coincidence to the timing of this realization.}
Put simply, the more tables with numbers there are, the less a table with numbers means, and the less it can be trusted.

\subsection{Number as Vice}
Its failure to really adhere to the positivist ethos does not absolve computational linguistics from having adopted it in the first place.
The idealization of science as an entity far and above subjective human reference provides the grounds for its disconnect from social context; there's no reflection on its production and consumption, the people involved in it and the people affected by it, or its effect on broader society and the world at large.
This detachment is reinforced by a techno-determinist narrative of ``progress'' moving of its own accord, which the scientist neither can influence, nor is responsible for~\cite{wyatt2008technological}.
Tables with numbers are the embodiment of techno-determinism.
The quest for experimental superiority (\ie{} ``progress'') is perceived as a self-efficient treadmill that continues on, regardless of who walks it -- there's no challenging the pace.

Setting off from a different axiomatization of scientific truth enables different inference paths.
Aiming for a computational linguistics that is attuned to the world allows us to challenge this particular interpretation of progress -- not just for its lack of scientific merit, but more importantly for its active role in perpetuating and amplifying environmental and social harm.
We build on this second perspective in the following sections.

\section{Resource Exhaustion}\label{sec:environmental}
Tables with numbers are the produce of large-scale experiments and extensive comparative evaluations.
They are displayed primarily to demonstrate numerically quantifiable superiority over competing cutting-edge technologies.
Whatever the epistemic merit of this practice, there is no denying its environmental cost.
As the field is witnessing a constant influx of progressively larger models, each vying for supremacy over increasingly more challenging benchmarks,
tables are growing in both size and count, and the numbers within are getting more resource-intensive by the day.
As a result, the environmental footprint of contemporary research is spiraling out of control~\cite{strubell2019energy}.

\subsection{No NLP to Be Done on a Dead Planet}
The point has resonated with the ecological sensibilities of the community, prompting a number of responses to the issue.
By now, these have come to coalesce into a niche of their own, united under the common banner of a so-called ``green AI''~\cite{schwartz2020green}.
So far, most of this green literature has gravitated around two thematic pillars~\cite{verdecchia2023systematic}.
The first involves matters of high-level policy: promoting greener models, raising awareness, stamping algorithms and models with eco-labels, \etc{}
The second involves matters of low-level practice: truncating or quantizing models, optimizing resource utilization, improving performance-to-emission ratios, \etc{}
While both are valuable research avenues, neither really addresses the essence of the problem: the benchmarking practice itself.
Indeed, ecological condemnations of the current \textit{modus operandi} are rare and far between~\cite{brevini2020black,brevini2021ai,brevini2022dispelling,heilinger2024beware}.

In this case, failing to note the obvious is not (just) a problem of deductive inadequacy; the omission is actually a take in disguise.
An ideological child of techno-determinism, on the one hand, and eco-modernism, on the other, it implicitly but loudly proclaims that there is no standing in the way of progress -- yet \textit{good} progress can save the world!
The incompatibility of these two positions is glaring.
There's little point debating the inherent benevolence of a progress that we cannot contest or control.
That said, there is no need to shy away from connecting the dots either. 
Tables with numbers are killing the planet; computational linguistics can never truly be green as long as it remains attached to them.
The ecologically responsible course of action is not to alleviate the effects -- it is to dismantle the cause.

\section{Institutional Bias \& Privilege}\label{sec:societal}
Besides environmental concerns, keeping up with contemporary research trends comes at a (literal) heavy price.
As the cost of the ``\textit{state of the art}'' explodes at a super-exponential rate~\cite[\ia]{sharir2020cost,epoch2023aitrends,perrault2024artificial}, the severe budget inequalities in higher education become further pronounced~\cite{o2016academic,goyes2023rich}, and the minimum requirements for scientific relevance becoming prohibitively high for smaller and lesser-funded institutions to acquire and maintain~\cite{ahmed2020democratization}.
Consequently, a few dominant institutions get to consolidate their competitive advantage by effectively gatekeeping the means necessary to conduct exactly the kind of research that is perceived as groundbreaking and impactful~\cite{munch2014academic}. 
This is problematic on multiple levels.

\begin{figure}[t]
    \centering
    \includegraphics[clip, trim=0.5cm 0 0 0,width=0.49\textwidth]{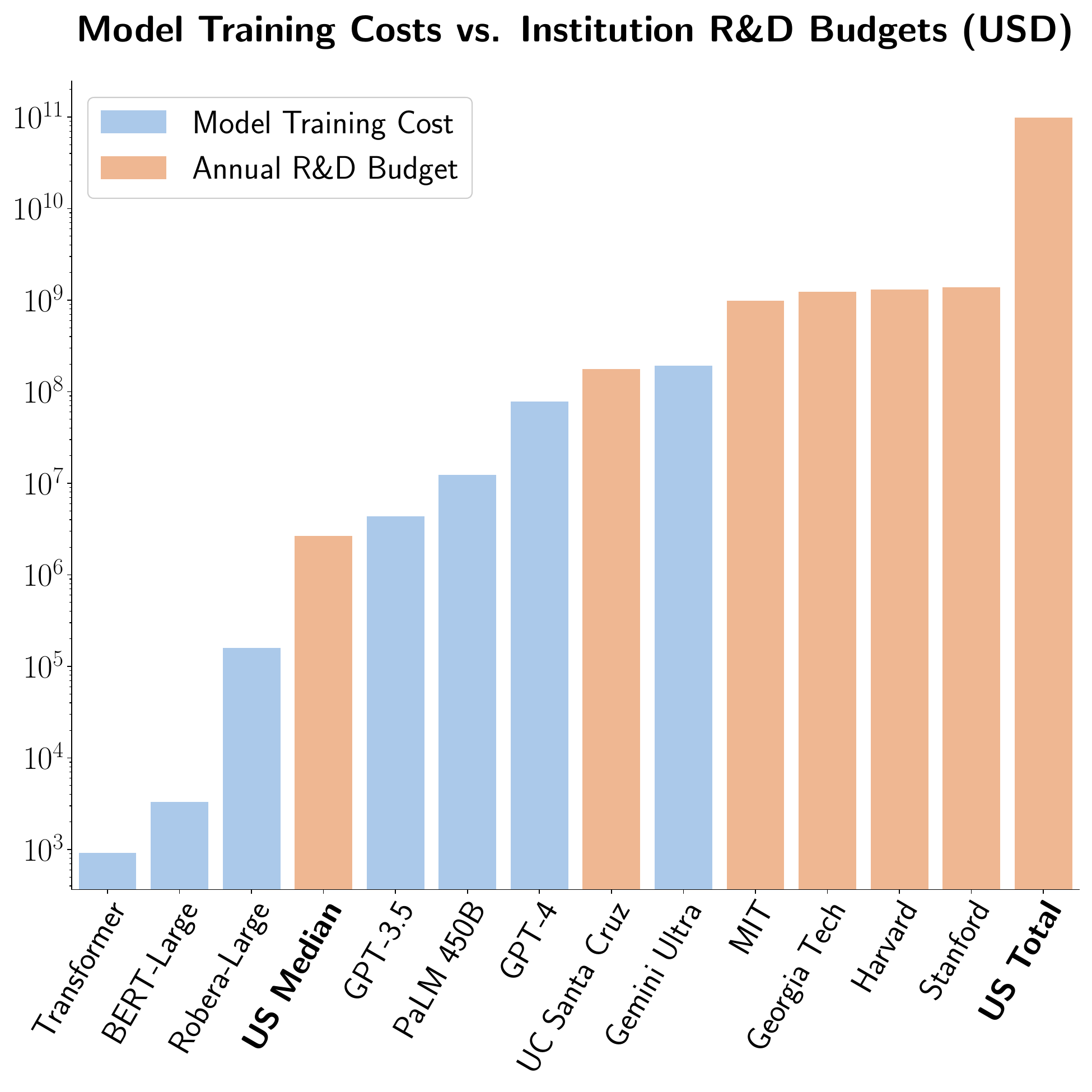}
    \caption{Contemporary model training costs compared to the total annual R\&D budgets of select U.S. institutions. The cost of training a large model is comparable to the budget of a university in the top 15th percentile, which is two orders of magnitude larger than the median budget. Budget data sourced from the 2022 report by the US National Center for Science and Engineering Statistics\textsuperscript{a}. Model cost estimates from~\citet{epoch2023aitrends}.}
    \begin{flushleft}
    \small\textsuperscript{a} \url{https://ncsesdata.nsf.gov/profiles/}    
    \end{flushleft}
    \label{fig:budgets}
\end{figure}

\subsection{Science of the Few}
To begin with, the insurmountable entry barrier perpetuates and exacerbates a cycle of entrenched privilege, where only a few voices retain access to the platforms of expression.
This disparity translates the lack of diversity in \textit{what} research is done to a lack of diversity in \textit{who} gets to do it~\cite{ahmed2020democratization,perrault2024artificial}.
For those favored, the cycle is no easier to break. 
The current status quo presents a very alluring prospect: a research recipe that is universally recognized as superior, and that only few have the ingredients necessary to implement. 
Opting out is not just a matter of critical reflection -- it is actually harmful to one's own interests (as measured in publications, citation counts, employment opportunities, \etc{}). 
Beyond the individual, the same dynamics appear at the institutional scale.
Steering a unit away from the competition for experimental superiority and towards niche research means condemning it into academic obscurity and irrelevance; both too easy to mistake for incompetence.
This further disincentives scientific plurality, placing the field on a convergent path toward a cuisine consisting of a single dish, served in microvariations by a handful of cooks, all of them too cautious not to stray lest they lose their status and privilege.

Effect being all too easy to mistake for cause, a few institutions have by now come to be lauded as hubs of research pioneers, their output singled out and preemptively lauded on the basis of origin alone~\cite{rigney2010matthew,brennen2019industry}.
Privileged individuals are granted undue influence over the field's trajectory, effectively getting to dictate both \textit{what questions to ask} (\eg{} which datasets to tackle), and \textit{where to look for the answers} (\eg{} which models to adopt).
This concentration of technical and scientific authority creates clearly delineated points of vulnerability for the field.
Alternative viewpoints and methodologies are at an increased risk of being left unnoticed or becoming squelched, suppressing innovation and inducing inertia.
Worse yet, it allows for the biases, norms and opinions of a few dominant actors to be perpetuated unhindered, now disguised as universal and irrefutable truths characterizing the entire discipline.

\subsection{Science for the Few}
This last issue is exacerbated exactly by the inherent narrowness of these biases, norms and opinions. 
Prestigious (read: \textit{wealthy}) institutions are neither evenly distributed across geographic regions, nor equally accessible across social, cultural, ethnic and economic backgrounds. 
As such, the perspectives and priorities they represent are inevitably skewed towards certain demographics, fostering homogenization at the expense of further marginalizing under-represented groups and identities~\cite[\ia]{amsler2012university,shamash2018re,field2021survey,talat-etal-2022-reap,hershcovich2022challenges,bender2024power,perrault2024artificial}. 
On the premise that cultural diversity is indeed worth nurturing and preserving~\cite{diversity}, the absence of plurality caused by this delegation of scientific and technological authority is bad -- for any scientific field.
For a field like computational linguistics in particular, it is \textit{catastrophic}.
Allowing research agendas to be shaped by a handful of actors endorses hegemonialism: not just technological and scientific, but importantly also cultural and linguistic.
Trending terms like ``natural language understanding'' carefully conceal the assumptions made on \textit{which} languages are actually worth understanding -- or what \textit{understanding} means, for that matter~\cite{bender2021dangers}.
The perspective that chasing after benchmarks and competing for the top spots in scoreboards carries some sort of inherent value to the computational study of language becomes immediately exposed as biased and flawed upon noticing that the majority of benchmarks and scoreboards pertain only to a minuscule fragment of the globe's peoples~\cite{joshi2020state,ruder2022statemultilingualai}.

Finally, a disproportionate allocation of resources creates the necessary preconditions for scientific tokenism.
Technological abundance for the few is indistinguishable from technological sparsity for the many.
The surging pressure for inclusivity is temptingly easy to relieve, either by reducing the bar when it comes to work in under-represented languages and cultures, or by ``allowing'' it to coexist along the mainstream as a secondary, self-referential niche.
And while this might indeed expedite its progress or increase its visibility, it carries the risk of negatively impacting its (perceived) quality, further cementing the gap between center and periphery worlds -- in terms of language, culture and research alike.

\subsection{From Inequality to Alienation}
Along the same lines, in monopolizing the resources essential for ``frontier'' research, ``world-class'' institutions gain a competitive edge in attracting highly sought-after global talent.
Predictably, transnational academic mobility flows along research capacity gradients shaped by global wealth inequalities~\cite{bilecen2017introduction}.
The exclusivity of ``frontier'' research turns academic mobility into a violent dilemma: move, or (academically) perish.
Built on this premise, ``frontier'' research cannot but carry a commodified and socially charged undertone~\cite{stein2017internationalization}.

Two orthogonal aspects of this perspective share a single common effect.
First, the same process that accelerates well-funded and globally competitive research decelerates regional institutions and projects by starving them of (yet) another precious resource: talent~\cite[\ia]{auriol2013careers,van2015international}.
Second, the inherently globalized nature of benchmarking and its constructed significance means that researchers employed abroad are predominantly engaged with work far detached from their own cultural and linguistic heritage.
The mirror image of an international researcher pushing the boundaries of ``cutting-edge'' research is an expatriated researcher not getting their own mother tongue up to speed with that very same research.
This reveals benchmarking as a driver for scientific assimilation, which turns linguistic coverage into a matter of institutionalized charity -- left to the discretion of exactly those fueling and benefiting from its absence.

\section{Science \& Profit}\label{sec:monetary}
Albeit alarming, institutional bias is to some extent mitigated by a common (if subjective and vague) promise of scientific integrity, a culture of transparency and openness, a shared strive for intellectual inquiry, and the self-regulatory effect of the occasionally functional peer-reviewing system~\cite{rogers2020can,Rogers_2020_reviewing-models}.
However, as the race for experimental superiority intensifies and turns increasingly exclusive, each new milestone gains greater appeal.
Beyond signaling intellectual achievement or academic accomplishment, this appeal extends to the material plane.
There, leading the benchmark race translates to a tangible competitive edge in commercial (and/or state) applications.
The allure of such an edge has been persistently attracting profit-driven entities into the computational linguistics ecosystem.
Over the span of a decade, these entities have evolved from circumstantial players to dominant figureheads.
For such entities, \textit{none} of the safeguards above hold.
This reality poses an existential threat for the field; a threat which nonetheless remains largely unaddressed.

\subsection{Stand on the Shoulders of (Tech) Giants}
The current state of affairs can be traced to a historical affinity between computational linguistics and machine learning~\cite{manning2015computational}.
Such an affinity is hardly surprising.
Language poses challenges at a variety of modalities and difficulty scales, enacting a boundless source of benchmarks for machine learning models.
Conversely, models and techniques developed for language-related tasks have frequently demonstrated their versatility as general-purpose machine learning tools, making their way to distant or even unrelated disciplines.
Until recently, this reciprocal relationship has been beneficial to both fields.
In the last few years, however, and as the pace of progress in machine learning has been consistently exceeding expectations, computational linguistics has lost its primacy, becoming increasingly dependent on imported expertise.
This trend is reflected in the silent but perfectly evident shift of the field's main inquiries, which have gradually moved from the computational study of language to an evaluation arena for application-oriented machine learning.
And even though this transition might disappoint or alienate some, there is not much inherently wrong about it; after all, it is not uncommon for a research field to retroactively change direction, or even be altogether absorbed or subsumed by another.
What \textit{is} problematic in the present context is the nature of the subsumer.

The main pathology of machine learning, having become synonymous with AI, is none other than its public and commercial appeal.
The commercialization of science demands tangible advantages against competitors: the product is easier to sell when it's visibly and quantitatively better than alternatives.
The success of this commercialization depends largely on ``\textit{wow!}'' factors: publicity stunts, catchy claims, and a degree of speculative futurism~\cite{funk2019s}.
For the global actors invested in the AI race, the concept of performance is thus of prime interest~\cite{bourne2024ai}.
Current technology dictates one base ingredient as the necessary and sufficient condition for performance: scale~\cite{epoch2023aitrends}.
And so, we get once more caught up in a vicious cycle.
As profit requires performance, profit requires scale and scale requires budget, a positive feedback loop ensures the growth of a handful of tech giants -- at a rate far exceeding that of even the wealthiest research institution.
And as performance just so happens to be our currency of choice when quantifying scientific advancement~\cite{birhane2022values}, machine learning research becomes \textit{de facto} dominated by exactly these giants~\cite{perrault2024artificial,de2024shift}.
This elevates the resource allotment problems discussed earlier to an altogether different scale: what's at stake now is not just equal and fair access to an equal and fair science, but rather the very idea of independent scientific inquiry~\cite{abdalla2021grey,jurowetzki2021privatization}.

\begin{figure}
    \centering
    \includegraphics[width=.49\textwidth,trim=0cm 0cm 0cm 0cm,clip=]{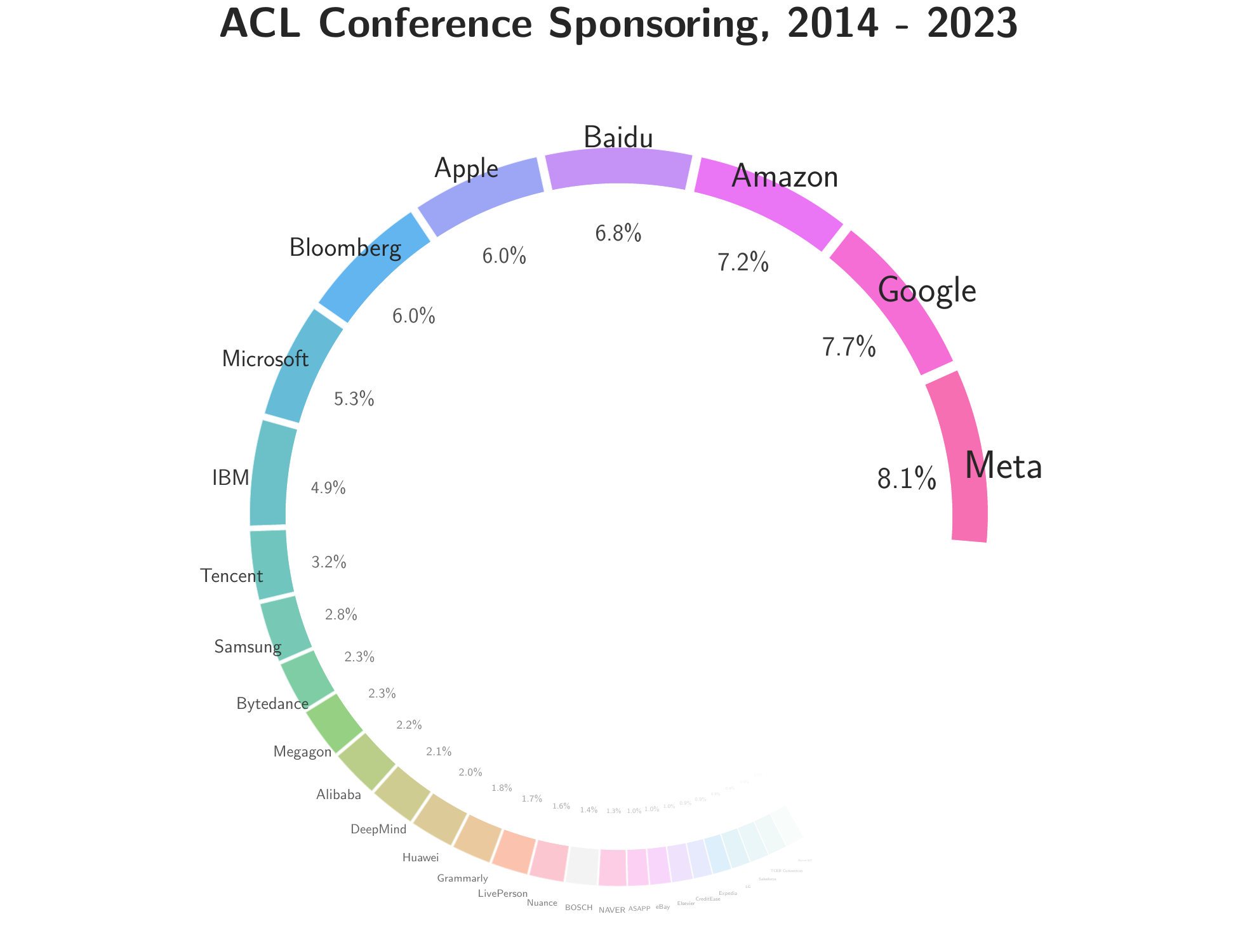}
    \caption{Major sponsors of the main ACL conferences over the last 10 years. To convert tiered participation counts to contributions, we assign a weight of 1 to the year's top tier, and divide the weight of each consecutive sponsorship tier by 2. The treasurer of the ACL did not respond to our request for accurate donation figures.}
    \label{fig:sponsors}
    \vspace{1em}
    \includegraphics[width=.49\textwidth]{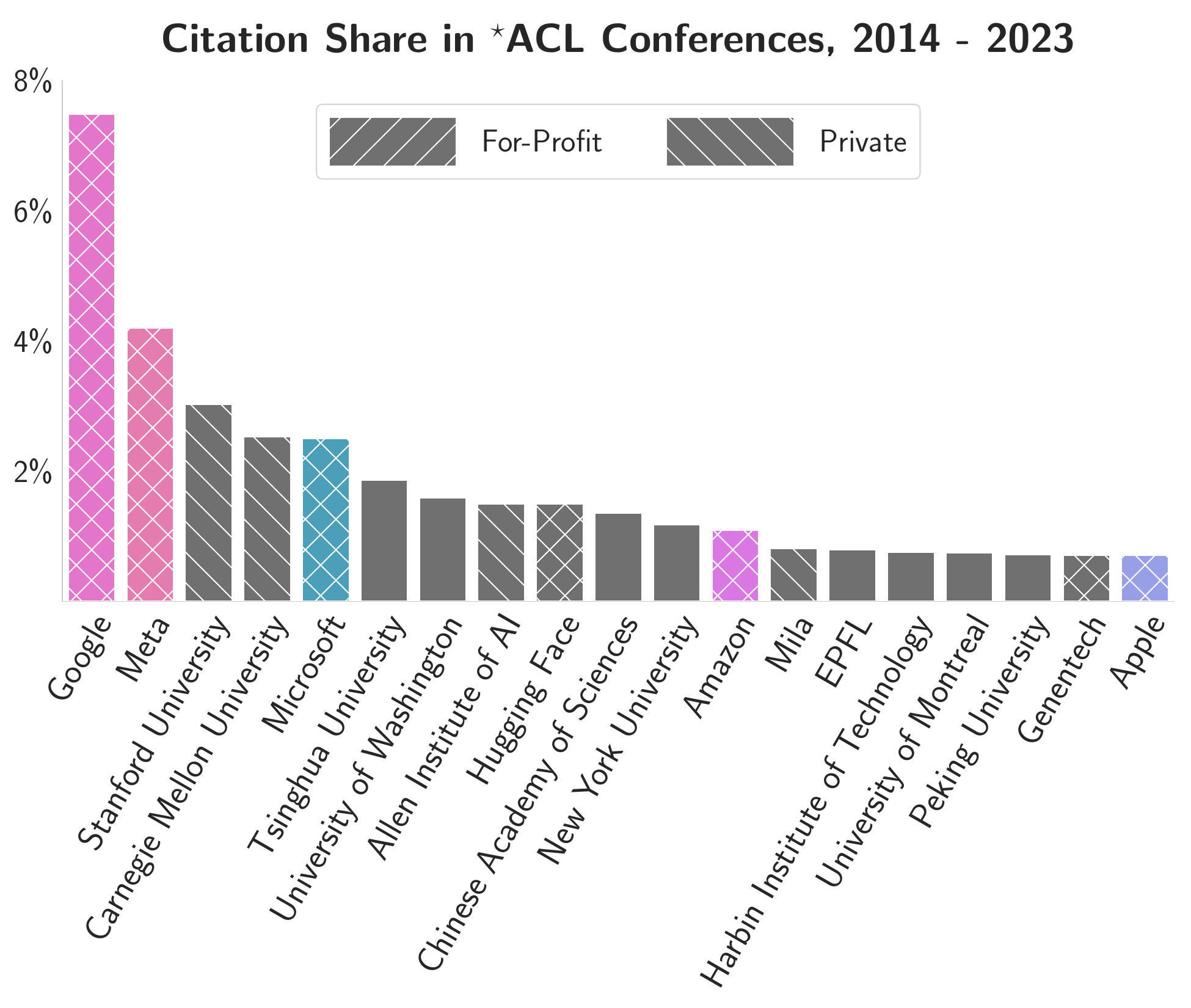}
    \caption{Citation share by organization in ${}^\star$\!ACL conferences over the last 10 years. The 19 organizations listed amount for approximately one third of the total citations during this period. 
    We associate (i) papers to authors, by parsing the ACL bibliography file, (ii) authors to affiliations, by crawling \href{https://scholar.google.com/}{\texttt{google scholar}} with \href{https://github.com/scholarly-python-package/scholarly}{\texttt{scholarly}}, and (iii) papers to publication counts, using \href{https://www.zotero.org}{\texttt{Zotero}} and the \href{https://www.github.com/FrLars21/ZoteroCitationCountsManager}{\texttt{ZoteroCitationCountsManager}} plugin.
    We collapse affiliations to organizations (\ie{} remove job titles and departments) by instructing \texttt{mistral-7B}~\cite{jiang2023mistral}.
    We compose and aggregate over the above to produce a map from organizations to citation counts, disregarding organizations with less than 5 citations as likely parsing errors.
    The result is imperfect: there are multiple sources of error, and affiliations at retrieval time are likely to differ from those at publication time."
    Nonetheless, it paints a sufficiently clear picture of which organizations are exerting the most influence in the field, and what the extent of this influence is.
    }
\end{figure}

\subsection{Research (and Development)}
In practice, as long as computational linguistics research remains results-oriented, reliance on technology and infrastructure provisioned by tech giants is a nonchoice -- there is, after all, \textit{no one else} to provision them from~\cite{abdalla2023elephant,ferrari2023neural}.
One might argue that such an arrangement is not without merits.
The narrative would usually be that putting corporate technology into the scientific spotlight facilitates the assessment of its risks and potentials, promoting accountability through transparency.
Conversely, integrating corporate resources into academia accelerates the actualization of research and increases its impact: rough prototypes turn into concrete tools, ensuring that scientific advancements reach the public domain faster.
But such a narrative depends on, and in fact presupposes, an alignment between scientific and commercial agendas.
The implication is that the pursuit of knowledge becomes conditional on its compatibility with the interests and capabilities of big tech, \ie{} the very same actors academia was supposed to scrutinize in the first place.

The conflict of interest is immediately apparent.
The overwhelming power asymmetry between big tech and academia (be it big or otherwise) erodes any potential merits that could ever be argued for.
Under the present conditions, the scientific spotlight can no longer be critical or investigative.
It devolves, instead, to a campaigning stage, a ticketed arena for the colossi to lift their veils and clash in public, and a marketplace for them to display their latest wares and recruit new talent.
Corporate resources do not ``spill over'', nor do they ``trickle down'' -- they are rationed; a means of scientific coercion~\cite{noble1979america,moore2011science,phan2022economies}.
Corporate interests do not actualize knowledge -- they predate, appropriate and monetize it~\cite{rikap2022big}.
Ideas that survive the ecosystem's selection process do not turn into socially relevant tools -- they turn into economically viable products~\cite{dale2019nlp,klinger2020narrowing,luitse2021great}.
Scientific involvement itself degrades into a ``networking filter'': an inconvenient but unavoidable stepping stone towards a high-stakes career in tech~\cite{ahmed2023growing,gofman2024artificial}.
The researcher becomes a glorified spokesperson for big tech, a consumer of their infrastructure, a public advocate of their science, a safety net between them and the public -- an eager and dispensable part of their production pipeline.

The extent and degree of the infiltration have become impossible to ignore.
We are on the verge of a corporate takeover, legitimized by an acquired taste for big datasets, big models and big numbers.
Put simply, we have been voluntarily handing the field over to an industry we are realistically incapable of challenging, let alone regulating.

\subsection{(The Irrelevance of) Corporate Ethics}
As of late, the community's growing awareness~\cite{michael2023nlp} of these developments and their public ramifications has spurred numerous works on so-called ``AI ethics''.
The conversation is heavily skewed by well-documented lobbying efforts and a broader ethics-washing campaign aimed at soothing public concern and deterring regulatory oversight.
The ``debate'' often revolves around virtue signaling gestures, assertions of corporate responsibility (or accusations of its absence), suggestions for self-regulatory accountability guidelines, techno-positive musings of an all-inclusive tomorrow, ``critical'' perspectives from within, vague calls for a misconstrued ``democratization'', and the like.
In their majority, these works range from malicious manipulation at worst, to harmful diversions at best~\cite[\ia]{ochigame2019invention,benkler2019don,slee2020incompatible, hagendorff2020ethics,phan2022economies,seele2022greenwashing,himmelreich2023against}.

This premeditated and narrow notion of ethics subtly chooses to ignore the possibility of us actually reclaiming control of the scientific discourse.
Besides negotiating matters of representation and inclusion, bias aversion, model explainability, linguistic diversity, open-sourcing, carbon impact, \etc{} as they arise within the current environment, we have a far more fundamental series of questions to be confronted with.
Are we assuming that big tech, running rampant on the field's collective advancements, will (or even can) ever align their agenda with the public's interests?
Do we trust them with upholding the values of scientific integrity and technological accountability?
Are we at peace with the prospect of a privatized and application-centric future for computational linguistics, removed from the world, her people and their needs?
If the answer to the above is no, how can we justify our implicit yet unwavering support and commitment to big tech's cause throughout the last decade?
Why are we so susceptible to their influence, so eager to adopt their values and principles, so tolerant of their technologically exclusionary practices?
Ultimately, what benefits do \textit{we} get to derive from contributing to \textit{their} endeavors -- and at what cost?

\section{Ways Ahead}
The paradigm shift advocated for might seem radical or untenable.
In reality, it is neither.
The epistemic rewiring it calls for does not require any kind of organized collective action; it can be advanced with a modest adjustment in research consumption and production attitudes at the individual level.
To effect this change, all we must do is reevaluate how we engage with research. 

As \textit{readers}, we need to stop allowing ourselves to be dazzled by big numbers.
We have to ask ourselves what their utility and cost is; but also who gets to benefit, and who bears the expense.

As \textit{authors}, we have to be conscious of our research goals and practices. 
We ought to look beyond numbers and benchmarks and focus on what questions our research really answers.
We must challenge the notion of science as a competition, and scorn endeavors that depend solely on experimental superiority to be deemed successful.
We must be mindful of the resources we use and their accessibility, but also of the artifacts we produce and their inclusivity. 
Above all, we must be upfront and self-critical about these matters, and we should hold our peers to the same standards.

As \textit{colleagues} and \textit{advisors}, it is our responsibility to be vocal about the issues in our field, and to find ways to promote better principles and ideals.

As \textit{reviewers}, we should each recognize our respective academic privileges, and be cautious in our technical demands; not everyone has access to the same number of GPUs.
Conversely, we should not be intimidated by big tables and bold face fonts; we need to be critical of the research we are exposed to, and call out opaque methodologies, exclusionary practices and useless flourishes.
Finally, our exclusive access to the reviewing process means it is our own duty to monitor it;  each one of us has a role in identifying and confronting poor practices.

\section{Conclusion}
We discussed tables with numbers, and related them to several issues that affect contemporary computational linguistics research.
We argued that the focus on experimental superiority has shifted research priorities towards technical optimization, at the expense of theoretical depth and societal context.
This has led to an inflationary effect on the epistemic value of experimental results, rendering them (and, by extension, the field itself), increasingly meaningless. 
We explained how the pressure for experimental superiority, while advancing technology, has fostered environmental degradation, institutional biases, and the commodification of research. 
To address these issues, we urge the field to critically reassess its methodologies, and prioritize a more holistic and socially responsible approach to scientific inquiry, balancing technical achievements with ethical and environmental considerations. 
This shift is essential for ensuring that advancements in computational linguistics positively contribute to scientific knowledge, societal well-being and environmental sustainability.\nolinebreak

\newpage

\section*{Limitations}
We tried to substantiate our claims with (references to) empirical evidence and contemporary critical perspectives. 
Nonetheless, this paper is first and foremost an opinion piece; the ideas presented are the product of subjective and ideologically signed mental processes.
For a reader that ascribes to the epistemic foundations of positivism, this is an argumentative weakness.
For us, it is a strength. 
We acknowledge our biases and limitations, and welcome critiques from all angles; a broader discussion on the field's epistemic culture is exactly what our work hopes to instigate.

Our analysis is by no means exhaustive, especially considering the complexity of the subject matter.
There are several experiments we would have hoped to carry out to quantify some of our claims, but we failed to bring to fruition.
We explicitly mention them here for the sake of clarity and transparency, and to bring them to the attention of other interested parties:
\begin{itemize}
    \item A paper-wise computational cost estimation would allow a quantification of the budgetary entry barrier to modern research. Overlaid with citation counts, this would allow answering whether the most impactful papers are really just the most expensive ones.
    \item A longitudinal topic modeling analysis could reveal the narrowing of research topics and methodologies over the last decade. Combined with an evolutionary analysis of writing norms (e.g., paper structure), this would allow us to correlate homogenization of tone with the loss of content diversity.
    \item A citation network analysis would make the field's citation dynamics more evident. Geographically mapping out central citation hubs could reveal underlying global research networks and allow for correlation with global economic indicators.
\end{itemize}

\section*{Acknowledgments}
Stergios Chatzikyriakidis gratefully acknowledges funding from the Special Account for Research Funding of the Technical University of Crete (grant number: 11218), as well as funding from the TALOS-AI4SSH ERA Chair in Artificial Intelligence for Humanities and Social Sciences grant (grant agreement: 101087269). 

\bibliography{custom}

\end{document}